\documentclass{article}

\PassOptionsToPackage{numbers}{natbib}

\usepackage[final]{nips_2017}

\usepackage[utf8]{inputenc} 
\usepackage[T1]{fontenc}    
\usepackage{hyperref}       
\usepackage{url}            
\usepackage{booktabs}       
\usepackage{amsfonts}       
\usepackage{nicefrac}       
\usepackage{microtype}      

\usepackage{algorithm}
\usepackage{algorithmic}
\usepackage{xcolor}
\usepackage{amsmath}
\usepackage{amsthm}
\usepackage{amssymb}
\usepackage{tikz}
\usepackage{etoolbox}
\usepackage[caption=false]{subfig}
\usepackage{multicol}
\usepackage{comment}

\usetikzlibrary{positioning}
\usetikzlibrary{arrows}

\newtheorem{theorem}{Theorem}

\newcommand{\eff}{\mathnormal{f}}

\title{Gauging Variational Inference}

\author{
Sungsoo Ahn\footnotemark[1] \qquad Michael Chertkov\footnotemark[2] \qquad Jinwoo Shin\footnotemark[1]\\
\footnotemark[1] School of Electrical Engineering,\\
Korea Advanced Institute of Science and Technology, Daejeon, Korea\\
\footnotemark[2] $^{1}$ Theoretical Division, T-4 \& Center for Nonlinear Studies,\\
Los Alamos National Laboratory, Los Alamos, NM 87545, USA,\\
\footnotemark[2] $^{2}$Skolkovo Institute of Science and Technology, 143026 Moscow, Russia\\
\footnotemark[1] \texttt{\{sungsoo.ahn, jinwoos\}@kaist.ac.kr}\qquad
\footnotemark[2] \texttt{chertkov@lanl.gov}\\}

\begin{document}

\maketitle

\begin{abstract}
Computing partition function is the most important statistical inference task arising in applications of Graphical Models (GM). Since it is computationally intractable, approximate  methods have been used to resolve the issue in practice, where mean-field  (MF) and belief propagation (BP) are arguably the most popular and successful approaches of a variational type. In this paper, we propose two new variational schemes, coined Gauged-MF (G-MF) and Gauged-BP (G-BP), improving MF and BP, respectively. Both provide lower bounds for the partition function by utilizing the so-called gauge transformation which modifies factors of GM while keeping the partition function invariant. Moreover, we prove that both G-MF and G-BP are exact for GMs with a single loop of a special structure, even though the bare MF and BP perform badly in this case. Our extensive experiments, on complete GMs of relatively small size and on large GM (up-to 300 variables) confirm that the newly proposed algorithms outperform and generalize MF and BP.
\end{abstract}
\section{Introduction}
Graphical Models (GM) express
factorization of the joint multivariate probability distributions
in statistics via a graph of relations between variables.
The concept of GM has been developed and/or used successfully in information theory \citep{gallager1962low,kschischang1998iterative},
physics \citep{35Bet,36Pei,87MPZ,parisi1988statistical,09MM}, artificial intelligence \citep{pearl2014probabilistic}, and machine learning \citep{jordan1998learning,freeman2000learning}.
Of many inference problems one can formulate using a GM,
computing the partition function (normalization),
or equivalently computing marginal probability distributions,
is the most important and universal inference task of interest.
However, this paradigmatic problem
is also known to be computationally intractable in general, i.e.,
it is \#P-hard even to approximate
\citep{jerrum1993polynomial}.

The Markov chain monte carlo (MCMC) \cite{alpaydin2014introduction} is 
a classical approach addressing the inference task, but
it typically suffers from exponentially slow mixing or large variance.
Variational inference is an approach stating the inference task
as an optimization.
Hence, it does not have such issues of MCMC and is often more favorable. 
The mean-field (MF) \citep{parisi1988statistical} 
and belief propagation (BP) \citep{pearl1982reverend} are
arguably the most popular algorithms of the variational type. 
They are distributed, fast and overall very successful in practical applications even though they are heuristics lacking systematic error control.  This 
has motivated researchers to seek for methods with some guarantees, e.g., providing lower bounds
\citep{liu2012negative, ermon2012density}
and upper bounds
\citep{wainwright2005new,liu2011bounding, ermon2012density} for 
the partition function of GM.

In another line of research, which this paper extends and contributes, the so-called re-parametrizations \citep{03WJW}, gauge transformations (GT) \citep{06CCa,06CCb} 
and holographic transformations \citep{valiant2008holographic,al2011normal} were explored. 
This class of distinct, but related, transformations consist in modifying a GM by changing factors, associated with elements of the graph, continuously  such that the partition function stays the same/invariant.\footnote{See \cite{08JW,forney2011partition,Misha_notes} for discussions of relations between the aforementioned techniques.} In this paper, we choose to work with GT as the most general one among the three approaches. Once applied to a GM, it transforms the original partition function, defined as a weighted {series/sum} 
over states, to a new one, dependent on the choice of gauges. In particular, a fixed point of BP minimizes the so-called Bethe free energy \cite{05YFW}, and it can also be understood as an optimal GT \citep{06CCa,06CCb,chernyak2007loop,mori2015holographic}. Moreover, fixing GT in accordance with BP  results in the so-called loop series expression for the partition function \citep{06CCa,06CCb}. In this paper we generalize \citep{06CCa,06CCb} and explore a more general class of GT. This allows us to develop a new gauge-optimization approach which results in `better' variational inference 
schemes than one provided by MF, BP and other related methods.

\noindent {\bf Contribution.}
The main contribution of this paper consists in developing two novel 
variational methods, called Gauged-MF (G-MF) and Gauged-BP (G-BP), providing
lower bounds on the partition function of GM. 
While MF minimizes the (exact) Gibbs free energy under (reduced) product distributions,
G-MF does the same task by introducing an additional GT.
Due to the the additional degree of freedom in optimization,
G-MF improves the lower bound of the partition function provided by MF systematically. 
Similarly, G-BP generalizes BP, extending interpretation of the latter as an optimization of the Bethe free energy over GT \citep{06CCa,06CCb,chernyak2007loop,mori2015holographic}, 
by imposing additional constraints on GT forcing all the terms 
in the resulting series for the partition function to remain non-negative.
Thus, G-BP results in a provable lower bound for the partition function, while BP does not (except for log-supermodular models \citep{ruozzi2012bethe}).

We prove that both G-MF and G-BP are exact for GMs defined over  
single cycles, which we call `alternating cycle/loop', as well as over the line graphs. The alternative cycle case
is surprising as {it represents }
the simplest `counter-example' 
from \cite{weller2014understanding}, illustrating failures of MF and BP. 
For general GMs, we also establish that
G-MF 
{is better than, or at least as good as} 
G-BP.
However, 
we also develop novel error correction schemes for G-BP
such that the lower bound of the partition function provided by G-BP 
can also be improved systematically/sequentially, 
eventually outperforming G-MF 
on the expense of increasing computational complexity.
Such an error correction scheme has been studied for improving BP by considering 
the loop series consisting of positive and negative terms \cite{chertkov2008belief,ahn2016synthesis}.
Due to our design of G-BP, 
the corresponding series consists of only non-negative terms, which makes
much easier to improve the quality of G-BP systematically.

We further found that our newly proposed GT-based optimizations can be restated as smooth and unconstrained ones,
thus allowing efficient solutions via algorithms of a gradient descent type or 
any generic optimization solver such as IPOPT \citep{wachter2006implementation}.
We experiment with IPOPT 
on complete GMs of relatively small size and on large GM (up-to {300}  variables) of fixed degree, which
confirm that the newly proposed  algorithms outperform and generalize MF and BP.
Finally, note that all statements of the paper are made within the framework of the so-called Forney-style GMs \citep{forney2001codes} which is general as  it allows interactions beyond pair-wise (i.e., high-order GM) and  includes other/alternative GM formulations, 
such as factor graphs of \citep{03WJ}.
Our results using GT for variational inference provide
a refreshing angle for the important inference task,
and we believe it should be of broad interest in many applications involving GMs.

\section{Preliminaries}\label{sec:pre}
\subsection{Graphical model}
\label{subsec:GM}

\noindent {\bf Factor-graph model.}
Given (undirected) bipartite factor graph $G=(\mathcal{X},\mathcal{F},\mathcal{E})$,
a joint distribution of (binary) random variables
$x=[x_{v} \in\{0,1\}: v \in \mathcal{X}]$ is called a factor-graph
Graphical Model (GM) if it factorizes as follows:
\begin{equation*}
p(x) = \frac{1}{Z} \prod_{a \in \mathcal{F}}
\eff_{a}(x_{\partial a}),
\end{equation*}
where $\eff_a$ are some non-negative
functions called factor functions,
$\partial a \subseteq \mathcal{X}$
consists of nodes
neighboring factor $a$,
and the normalization constant
$Z:= \sum_{x\in \{0,1\}^{\mathcal{X}}}
  \prod_{a\in \mathcal{F}} \eff_{a}(x_{\partial a}),$
is called the partition function.
A factor-graph GM is called pair-wise if $|\partial a|\leq 2$ for all
$a \in \mathcal F$,
and high-order otherwise.
It is known that approximating the partition function
is \#P-hard even for pair-wise GMs in general \citep{jerrum1993polynomial}.

\noindent {\bf Forney-style model.}
In this paper, we primarily use the
Forney-style GM \citep{forney2001codes} instead of factor-graph GM.
Elementary random variables in the Forney-style GM
are associated with edges of an undirected graph,
$G = (\mathcal{V}, \mathcal{E})$.
Then the random vector,
$x=[x_{ab} \in \{0,1\}: \{a,b\} \in \mathcal{E}]$
is realized with the
probability distribution
\begin{equation}
p(x) =
\frac{1}{Z}
\prod_{a\in \mathcal{V}} \eff_{a}(x_{a}),
\label{Forney}
\end{equation}
where $x_{a}$ is associated with
set of edges neighboring node $a$, i.e.
$x_{a}= [x_{ab} ~:~ b\in \partial a]$
and
$Z := \sum_{x\in \{0,1\}^{\mathcal{E}}}
\prod_{a\in V} \eff_{a}(x_{a}).$
As argued in \cite{06CCa,06CCb},
the Forney-style GM constitutes a more universal and compact description of gauge transformations
without any restriction of generality, i.e.,
given
any factor-graph GM,
one can construct an equivalent Forney-style GM
(see the supplementary material).

\subsection{Mean-field and belief propagation}
\label{subsec:energy}
In this section, we introduce two most popular methods for approximating the partition function:
the mean-field and Bethe (i.e., belief propagation) approximation methods.
Given any (Forney-style) GM $p(x)$
defined as in \eqref{Forney} and
any distribution $q(x)$ over all variables,
the {\em Gibbs free energy} is defined as
\begin{equation}\label{eq:gibbs}
  F_{\text{Gibbs}}(q) :=\sum_{x \in \{0,1\}^{\mathcal{E}}} q(x) \log \frac{q(x)}{\prod_{a\in \mathcal{V}} \eff_{a}(x_{a})}.
\end{equation}
Then the partition function is derived according to $-\log Z = \min_{q}  F_{\text{Gibbs}}(q)$, where
the optimum is achieved at $q=p$, {e.g., see \citep{03WJ}.} 
{This optimization is over all valid probability distributions on the exponentially large space 
and obviously intractable.}


In the case of the {mean-field} (MF) approximation, we minimize the Gibbs free energy over 
a family of tractable probability distributions factorized into the following product:  
$q(x) = \prod_{\{a,b\}\in\mathcal{E}} q_{ab}(x_{ab})$,
where each independent $q_{ab}(x_{ab})$ is a proper probability distribution, 
{behaving as a (mean-field) proxy to the marginal of $q(x)$ over $x_{ab}$.}
By construction, the MF approximation provides a lower bound for $\log Z$. 
In the case of the {Bethe approximation}, the so-called
{\em Bethe free energy} approximates
the Gibbs free energy \citep{yedidia2001bethe}:
\begin{equation}
  \label{eq:bethe}
  F_{\text{Bethe}}(b) = \sum_{a\in\mathcal{V}}\sum_{x_{a}\in\{0,1\}^{\partial a}} b_{a}(x_{a})\log\frac{b_{a}(x_{a})}{\eff_{a}(x_{a})}
  - \sum_{\{a,b\}\in\mathcal{E}}\sum_{x_{ab} \in \{0,1\}}b_{ab}(x_{ab})\log b_{ab}(x_{ab}),
\end{equation}
where {\em beliefs} $b = [b_{a}, b_{ab}~:~ a \in \mathcal{V}, \{a,b\}\in \mathcal{E}]$
should satisfy following `consistency' constraints:
\begin{align*}
  &0\leq b_{a}, b_{ab}\leq 1, \quad \sum_{x_{ab}\in\{0,1\}} b_{a}(x_{ab}) = 1, \quad
  \sum_{x^{\prime}_{a}\backslash x_{ab} \in\{0,1\}^{\partial a}}b(x^{\prime}_{a}) = b(x_{ab})\quad \forall \{a,b\}\in\mathcal{E}.
\end{align*}
Here, $x^{\prime}_{a}\backslash x_{ab}$ denotes a vector with $x^{\prime}_{ab} = x_{ab}$ fixed 
{and} 
$\min_{b} F_{\text{Bethe}}(b)$ is the Bethe estimation for $-\log Z$.
The popular belief propagation (BP) distributed heuristics solves the optimization iteratively \citep{yedidia2001bethe}.
The Bethe approximation is exact over trees, i.e., $-\log Z = \min_{b}  F_{\text{Bethe}}(b)$.
However, in the case of a general loopy graph, the BP estimation lacks approximation guarantees. 
It is known, however,  that the result of BP-optimization lower bounds the log-partition function, $\log Z$, if the factors are log-supermodular \citep{ruozzi2012bethe}.

\subsection{Gauge transformation}\label{subsec:GT}
Gauge transformation (GT) \citep{06CCa,06CCb}
is a family of linear transformations of the factor functions in \eqref{Forney}
which leaves the
the partition function $Z$ 
invariant. 
It is defined with 
respect to the following set of  invertible $2\times 2$ matrices $G_{ab}$ for $\{a,b\}\in \mathcal{E}$, coined {\emph{gauges}}:
\begin{equation*}
  G_{ab}
  = \left[\begin{array}{ccc}
  G_{ab}(0,0) & G_{ab}(0,1)\\
  G_{ab}(1,0) & G_{ab}(1,1)\\
  \end{array}
  \right].
\end{equation*}
The GM, gauge transformed 
with respect to
$\mathcal{G} = [G_{ab}, G_{ba}~:\{a,b\} \in \mathcal E]$,
consists of factors expressed as:
\begin{equation*}
\eff_{a,\mathcal{G}}(x_{a})
= \sum_{{x}_{a}^\prime\in \{0,1\}^{\partial a}}\eff_{a}({x}_{a}^\prime)
\prod_{b \in \partial a} G_{ab}(x_{ab}, x^\prime_{ab}).
\end{equation*}
Here one treats independent $x_{ab}$ and $x_{ba}$ equivalently
for notational convenience, and $\{G_{ab},G_{ba}\}$ is a
conjugated pair of distinct matrices satisfying
the gauge constraint
$G_{ab}^\top G_{ba} = \mathbb{I}$,
where $\mathbb{I}$ is the identity matrix.
Then, one can prove invariance of the partition function under the transformation:
\begin{equation}\label{eq:invariant}
Z ~=~ \sum_{x\in \{0,1\}^{|\mathcal{E}|}}
\prod_{a\in \mathcal{V}} \eff_{a}(x_{a})
~=~ \sum_{x\in \{0,1\}^{|\mathcal{E}|}}
\prod_{a\in \mathcal{V}} \eff_{a,\mathcal{G}}(x_{a}).
\end{equation}
Consequently, GT results in the gauge transformed
distribution 
$p_{\mathcal{G}}(x) = \frac1Z \prod_{a\in \mathcal{V}}\eff_{a,\mathcal{G}}(x_{a}).$
Note that some components of  $p_{\mathcal{G}}(x)$ can be negative, in which case it is not a valid probability distribution. 

We remark
that the Bethe/BP approximation can be interpreted as 
a specific choice of GT \citep{06CCa,06CCb}. Indeed any fixed point of BP 
corresponds to a special set of gauges making an
arbitrarily picked configuration/state $x$ to be least
sensitive to the local variation of the gauge.
Formally,  the following non-convex optimization is known to be 
equivalent to the Bethe approximation:  
\begin{align}
    \operatornamewithlimits{\mbox{maximize}}_{\mathcal{G}}\quad
    &\sum_{a\in \mathcal{V}} \log\eff_{a,\mathcal{G}}(0,0,\dots)\notag\\
    \mbox{subject to}\quad
    &G_{ab}^{\top}G_{ba} = \mathbb{I}, \quad
    \forall ~ \{a,b\} \in \mathcal{E}, 
    \label{eq:betheopt}
\end{align}
and the set of BP-gauges correspond to stationary points of \eqref{eq:betheopt}, having
the objective as the respective Bethe free energy, i.e., $\sum_{a\in \mathcal{V}} \log\eff_{a,\mathcal{G}}(0,0,\dots) = -F_{\text{Bethe}}$.
\section{Gauge optimization for approximating partition functions}
\label{sec:alg}
{Now we are ready to describe two novel gauge optimization schemes (different from \eqref{eq:betheopt}) providing 
guaranteed lower bound approximations for $\log Z$.}
Our first GT scheme, coined Gauged-MF (G-MF), shall be considered as modifying and improving the MF approximation, while our second GT scheme, coined Gauged-BP (G-BP), modifies and improves the Bethe approximation {in a way that} 
it now provides a provable lower bound for $\log Z$,  while the bare BP 
does not have such guarantees. The G-BP scheme also allows further improvement (in terms of the output quality) on the expense of making underlying algorithm/computation more complex.

\subsection{Gauged mean-field}
\label{subsec:GTgibbs}

We first propose the following optimization inspired by, and also improving,  the MF approximation:
\begin{align}
    \operatornamewithlimits{\mbox{maximize}}_{q,\mathcal{G}}
    \quad
    &\sum_{a\in \mathcal{V}} \sum_{x_{a} \in \{0,1\}^{\partial a}} q_{a}(x_{a}) \log \eff_{a,\mathcal{G}}(x_{a}) - 
    \sum_{\{a,b\}\in \mathcal{E}}\sum_{x_{ab}\in\{0,1\}}q_{ab}(x_{ab})\log{q_{ab}(x_{ab})} \notag\\
    \mbox{subject to}\quad
    &G_{ab}^{\top}G_{ba} = \mathbb{I}, \quad
    \forall ~ \{a,b\} \in \mathcal{E}, \notag\\
    &\eff_{a,\mathcal{G}}(x_{a}) \geq 0, \quad
    \forall a\in \mathcal{V},~\forall x_{a}\in \{0,1\}^{\partial a},\notag\\
    & q(x)=\prod_{\{a,b\}\in{\cal E}} q_{ab}(x_{ab}),\quad q_a(x_a)=\prod_{b\in \partial a} q_{ab}(x_{ab}),\quad \forall a\in{\cal V}.
    \label{eq:gibbsopt}
\end{align} 
{Recall that the MF approximation optimizes the Gibbs free energy with respect to $q$ given the original GM, i.e. factors.
On the other hand, \eqref{eq:gibbsopt} jointly optimizes it over $q$ and $\mathcal G$.} 
Since the partition function of
the gauge transformed GM is equal to that of the original GM,
\eqref{eq:gibbsopt} also outputs a lower bound on the (original) partition function,
and always outperforms MF due to the additional degree of freedom in $\mathcal G$.
The non-negative constraints $\eff_{a,\mathcal{G}}(x_{a}) \geq 0$ for each factor
enforce that the gauge transformed GM results in a 
valid probability distribution (all components are non-negative).

\begin{algorithm}[ht!]
\caption{Gauged mean-field}
\label{alg:gibbsopt}
\begin{algorithmic}[1]
\STATE {\bf Input:} GM defined over graph $G=(\mathcal{V},\mathcal{E})$ with factors $\{\eff_{a}\}_{a\in\mathcal{V}}$. 
A sequence of decreasing barrier terms $\delta_{1}> \delta_{2}> \cdots> \delta_{T}>0$ (to handle extreme cases). 
\vspace{0.05in}
  \hrule
    \vspace{0.05in}
  \FOR{$t = 1,2,\cdots,T$}
\STATE {\bf Step A.} Update $q$ by solving the mean-field approximation, i.e., solve the following optimization:
\begin{align}
    \operatornamewithlimits{\mbox{maximize}}_{q}
    \quad &
    \sum_{a\in \mathcal{V}} \sum_{x_{a} \in \{0,1\}^{\partial a}} q_{a}(x_{a}) \log \eff_{a,\mathcal{G}}(x_{a}) - 
    \sum_{\{a,b\}\in \mathcal{E}}\sum_{x_{ab}\in\{0,1\}}q_{ ab}(x_{ab})\log{q_{ab}(x_{ab})}\notag\\
    \mbox{subject to}\quad
    & q(x)=\prod_{\{a,b\}\in{\cal E}} q_{ab}(x_{ab}),\quad q_a(x_a)=\prod_{b\in \partial a} q_{ab}(x_{ab}),\quad \forall a\in{\cal V}.\notag
\end{align}
\STATE {\bf Step B.} For factors with zero values, i.e. $q_{ab}(x_{ab}) = 0$, make perturbation by setting 
\begin{equation*}
\begin{split}
q_{ab}(x^{\prime}_{ab}) = \begin{cases} 
\delta_{t} \qquad &\text{if }x^\prime_{ab} = x_{ab}\\ 
1-\delta_{t} \qquad &\text{otherwise}.
\end{cases}
\end{split}
\end{equation*}
\STATE {\bf Step C.} Update $\mathcal{G}$ by solving the following optimization:
\begin{align*}
    \operatornamewithlimits{\mbox{maximize}}_{\mathcal{G}}
    \quad
    &\sum_{a\in V} \sum_{x \in \{0,1\}^{\mathcal{E}}} q(x) \log \prod_{a\in V} \eff_{a,\mathcal{G}}(x_{a}) \\
    \mbox{subject to}\quad
    &G_{ab}^{\top}G_{ba} = \mathbb{I}, \quad
    \forall ~ \{a,b\} \in \mathcal{E}. 
\end{align*}
  \ENDFOR
  \vspace{0.05in}
  \hrule
    \vspace{0.05in}
  \STATE {\bf Output:} Set of gauges $\mathcal{G}$ and product distribution $q$.
\end{algorithmic}
\end{algorithm}

To solve \eqref{eq:gibbsopt}, we propose a strategy, alternating between two optimizations,
formally stated in {Algorithm \ref{alg:gibbsopt}}. The alternation is between updating $q$, within {Step A}, and updating $\mathcal{G}$, within {Step C}. 
The optimization in {Step A} is simple as one can apply any solver
of the mean-field approximation.
On the other hand, {Step C} requires a new solver and, 
at the first glance, looks complicated 
due to nonlinear constraints. 
However,  the constraints can actually be eliminated. Indeed, one observes that
the non-negative constraint $\eff_{a,\mathcal{G}}(x_{a}) \geq 0$ 
is 
redundant,
because each term 
$q(x_{a})\log \eff_{a,\mathcal{G}}(x_{a})$ in the optimization objective already prevents
factors from getting close to zero, thus keeping them positive.
Equivalently, once current $\mathcal{G}$ satisfies the non-negative constraints,  
the objective, $q(x_{a})\log \eff_{a,\mathcal{G}}(x_{a})$, acts as a log-barrier forcing the constraints to be satisfied at the 
next step 
within an iterative optimization procedure. Furthermore, the gauge constraint, $G_{ab}^{\top}G_{ba} = \mathbb{I}$, 
can also be removed simply expressing one (of the two) 
{gauge} via another, e.g., $G_{ba}$ via $(G_{ab}^{\top})^{-1}$.  
Then, {Step C} can be 
resolved by any unconstrained iterative optimization method of a gradient descent type
or any generic optimization solver such as IPOPT \citep{wachter2006implementation}.
Next, the additional (intermediate) procedure {Step B} was considered to handle extreme cases when for some $\{a,b\}$, $q_{ab}(x_{ab})=0$ at the optimum. 
We resolve the singularity  
perturbing the distribution  by 
setting zero probabilities to a small value, $q_{ab}(x_{ab}) = \delta$ 
where $\delta > 0$ is sufficiently small.
In summary, it is straightforward to check that the
{Algorithm \ref{alg:gibbsopt}}
converges to a local optimum of \eqref{eq:gibbsopt}, similar to some other solvers developed for the mean-field and Bethe approximations.

We also provide an important class of GMs where 
the {Algorithm \ref{alg:gibbsopt}} provably outperforms both the 
MF and BP (Bethe) approximations.
Specifically, we prove that the optimization \eqref{eq:gibbsopt} 
is exact in the case when the graph 
is a line (which is a special case of a tree) and, 
somewhat surprisingly, 
a single loop/cycle
with odd number of factors
represented
by negative definite matrices.
In fact, the latter case is the so-called `alternating cycle' example
which was introduced in
\cite{weller2014understanding} as the simplest
loopy example where the MF and BP approximations perform quite badly.
Formally, we state the following theorem whose
proof is given 
in the supplementary material.
\begin{theorem}\label{thm:optcycle}
  For GM defined on any line graph or alternating cycle, the optimal objective of
  \eqref{eq:gibbsopt} is equal to the exact log partition function, i.e., $\log Z$.
\end{theorem}

\subsection{Gauged belief propagation}
\label{subsec:GTbethe}

We start discussion of the G-BP scheme by noticing that, according to \cite{chertkov2006loop}, the G-MF gauge optimization
\eqref{eq:gibbsopt} can be reduced to
the BP/Bethe gauge optimization \eqref{eq:betheopt} 
by eliminating the non-negative constraint $\eff_{a,\mathcal{G}}(x_{a}) \geq 0$ for each factor
and replacing 
the product distribution $q(x)$ by: 
\begin{equation}
\label{eq:qchoice}
\begin{split}
    q(x) = 
    \begin{cases}
    1 \qquad &\text{if } x = (0,0,\cdots),\\
    0 \qquad &\text{otherwise}.
    \end{cases}
\end{split}
\end{equation} 
Motivated by this observation, we propose 
the following G-BP optimization:
\begin{align}
    \operatornamewithlimits{\mbox{maximize}}_{\mathcal{G}}
    \quad
    &\sum_{a\in V}\log \eff_{a,\mathcal{G}}(0,0,\cdots)  \notag\\
    \mbox{subject to}\quad
    &G_{ab}^{\top}G_{ba} = \mathbb{I}, \quad
    \forall (a,b) \in \mathcal{E}, \notag \\
    &\eff_{a,\mathcal{G}}(x_{a}) \geq 0, \quad
    \forall a\in \mathcal{V},~\forall x_{a}\in \{0,1\}^{\partial a}.\label{eq:nnbetheopt}
\end{align}


The only difference between \eqref{eq:betheopt} and
\eqref{eq:nnbetheopt} is addition of the non-negative constraints for factors in \eqref{eq:nnbetheopt}. Hence, \eqref{eq:nnbetheopt} outputs a lower bound on the partition function,
while \eqref{eq:betheopt} can be larger or smaller then $\log Z$. 
It is also easy to verify that \eqref{eq:nnbetheopt} (for G-BP) is equivalent to \eqref{eq:gibbsopt} (for G-MF) with $q$ fixed to \eqref{eq:qchoice}.
Hence, we propose the algorithmic procedure for solving \eqref{eq:nnbetheopt},
formally described in {Algorithm \ref{alg:nnbetheopt}}, and
it should be viewed as a modification of {Algorithm \ref{alg:gibbsopt}} 
with $q$ replaced by \eqref{eq:qchoice} in {Step A}, also with a properly chosen log-barrier term in {Step C}. As we discussed for Algorithm \ref{alg:gibbsopt}, it is straightforward to verify that Algorithm \ref{alg:nnbetheopt} also converges to a local optimum of
\eqref{eq:nnbetheopt} and one can replace 
$G_{ba}$ by $(G_{ab}^{\top})^{-1}$ 
for each pair of the conjugated matrices 
in order to build a convergent gradient descent algorithmic implementation for the optimization. 

\begin{algorithm}[ht]
\caption{Gauged belief propagation}
\label{alg:nnbetheopt}
\begin{algorithmic}[1]
\STATE {\bf Input:} GM defined over graph $G=(\mathcal{V},\mathcal{E})$ with and factors $\{\eff_{a}\}_{a\in\mathcal{V}}$. A sequence of 
decreasing barrier terms $\delta_{1}> \delta_{2}> \cdots> \delta_{T}>0$. 
\vspace{0.05in}
  \hrule
    \vspace{0.05in}
  \FOR{$t = 1,2,\cdots$}
    \STATE  Update $\mathcal{G}$ by solving the following optimization:
    \begin{align*}
    \operatornamewithlimits{\mbox{maximize}}_{\mathcal{G}}
    \quad
    &\sum_{a\in V}\log \eff_{a,\mathcal{G}}(0,0,\cdots) + \delta_{t}\sum_{a\in V} \sum_{x \in \{0,1\}^{\mathcal{E}}} q(x) \log \prod_{a\in V} \eff_{a,\mathcal{G}}(x_{a}) \\
    \mbox{subject to}\quad
    &G_{ab}^{\top}G_{ba} = \mathbb{I}, \quad
    \forall ~ \{a,b\} \in \mathcal{E}. 
    \end{align*}
  \ENDFOR
  \vspace{0.05in}
  \hrule
    \vspace{0.05in}
  \STATE {\bf Output:} Set of gauges $\mathcal{G}$.
\end{algorithmic}
\end{algorithm}

Since fixing $q(x)$ eliminates the degree of freedom in \eqref{eq:gibbsopt},
G-BP should perform worse than G-MF, i.e., \eqref{eq:nnbetheopt} $\leq$ \eqref{eq:gibbsopt}.
However, 
G-BP is still meaningful due to the following reasons. 
First, Theorem \ref{thm:optcycle} still holds for \eqref{eq:nnbetheopt}, i.e.,
the optimal $q$ of \eqref{eq:gibbsopt} is achieved at \eqref{eq:qchoice}
for any line graph or alternating cycle (see the proof of the Theorem \ref{thm:optcycle} in the supplementary material).
More importantly,
G-BP can be corrected systematically.
At a high level, the ``error-correction" strategy consists 
in correcting the approximation error of \eqref{eq:nnbetheopt} sequentially 
while maintaining the desired lower bounding guarantee.
The key idea here is to
decompose the error of \eqref{eq:nnbetheopt} into
partition functions of multiple GMs,
and then repeatedly lower bound each partition function. 
Formally, we fix an arbitrary ordering of edges 
$e_{1},\cdots e_{|\mathcal{E}|}$
and define the corresponding GM for each $e_{i}$ as follows:
  $p(x) = \frac{1}{Z_{i}} \prod_{a \in \mathcal{V}} \eff_{a,\mathcal{G}}(x_{a})$
 for $x \in \mathcal{X}_{i}$,
where $Z_{i} := \sum_{x \in \mathcal{X}_{i}} \prod_{a \in \mathcal{V}}\eff_{a,\mathcal{G}}(x)$ and
\begin{equation*}
  \mathcal{X}_{i} := \{x~:~x_{e_{i}} = 1, x_{e_{j}} = 0, x_{e_{k}}\in \{0,1\}\quad\forall~ j, k,~\text{such that}~1 \leq j < i < k \leq |\mathcal{E}|\}.
  \end{equation*}
Namely, we consider {GMs from} sequential
conditioning {of} $x_{e_{1}}, \cdots, x_{e_{i}}$ in the gauge transformed GM. Next, recall that \eqref{eq:nnbetheopt}
maximizes and outputs a single configuration
$\prod_a \eff_{a,\mathcal{G}}(0,0,\cdots)$.
Then, since $\mathcal{X}_{i}\bigcap \mathcal{X}_{j} = \emptyset$ and
$\bigcup^{|\mathcal{E}|}_{i=1}\mathcal{X}_{i} = \{0,1\}^{\mathcal{E}} \backslash (0,0,\cdots)$,
the error of \eqref{eq:nnbetheopt} can be 
decomposed as follows:
\begin{equation}
\label{eq:errorseq}
  Z - \prod_a \eff_{a,\mathcal{G}}(0,0,\cdots) =
 \sum^{|\mathcal{E}|}_{i = 1}\sum_{x \in \mathcal{X}_{i}} \prod_{a \in \mathcal{V}}\eff_{a,\mathcal{G}}(x)
 = \sum^{|\mathcal{E}|}_{i=1}Z_{i},
\end{equation}
Now, one can run G-MF, G-BP 
or any other methods (e.g., MF) 
again to obtain a lower bound $\widehat{Z}_i$ of $Z_i$ for all $i$
and then output
$\prod_{a\in\mathcal{V}} \eff_{a,\mathcal{G}}(0,0,\cdots) + \sum^{|\mathcal{E}|}_{i=1}\widehat{Z}_{i}$.
However, such additional runs of optimization inevitably increase 
the overall complexity. 
Instead,  one can also pick a single term 
$\prod_{a} \eff_{a,\mathcal{G}}(x^{(i)}_{a})$
for $x^{(i)} = [x_{e_{i}} = 1, x_{e_{j}} = 0,~\forall~j \neq i]$ from $\mathcal{X}_{i}$, 
as a choice of $\widehat{Z}_{i}$ just after solving \eqref{eq:nnbetheopt} initially, and output
\begin{equation}\label{eq:fixerror1}
\prod_{a\in\mathcal{V}} \eff_{a,\mathcal{G}}(0,0,\cdots) + \sum^{|\mathcal{E}|}_{i=1}\eff_{a,\mathcal{G}}(x^{(i)}_{a}), 
\qquad x^{(i)} = [x_{e_{i}} = 1, x_{e_{j}} = 0,~\forall~j \neq i],
\end{equation}
as a better lower bound for $\log Z$ than $\prod_{a\in\mathcal{V}} \eff_{a,\mathcal{G}}(0,0,\cdots)$.
This choice is based on the intuition that 
configurations
partially different from $(0,0,\cdots)$ may be
significant too as they share
most of the same factor values
with the zero configuration maximized in \eqref{eq:nnbetheopt}. 
In fact, 
one can even choose more configurations {(partially different from $(0,0,\cdots)$)} by paying more complexity, 
which is always better as it brings the approximation closer to the 
true partition function. 
In our experiments, 
we consider {additional} 
configurations $\{x~:~ [x_{e_{i}} = 1, x_{e_{i^{\prime}}} = 1, x_{e_{j}} = 0,~\forall~i,i^{\prime} \neq j] 
\text{ for } i^{\prime} = i,\cdots |\mathcal{E}|\}$, 
i.e., output 
\begin{equation}\label{eq:fixerror2}
\prod_{a\in\mathcal{V}} \eff_{a,\mathcal{G}}(0,0,\cdots) + \sum^{|\mathcal{E}|}_{i=1}\sum^{|\mathcal{E}|}_{i^{\prime}=i}\eff_{a,\mathcal{G}}(x^{(i,i^{\prime})}_{a}), 
\quad x^{(i,i^{\prime})} = [x_{e_{i}} = 1, x_{e_{i^{\prime}}} = 1, x_{e_{j}} = 0,~\forall~ j \neq i,i^{\prime}],
\end{equation}
as a better lower bound of $\log Z$ than \eqref{eq:fixerror1}. 
\vspace{-0.1in}
\section{Experimental results}
\label{sec:experiments}

\begin{figure*}[t!]
    \vspace{-0.1in}
\centering
    \begin{minipage}{0.32\textwidth}
    \includegraphics[width=0.99\textwidth]{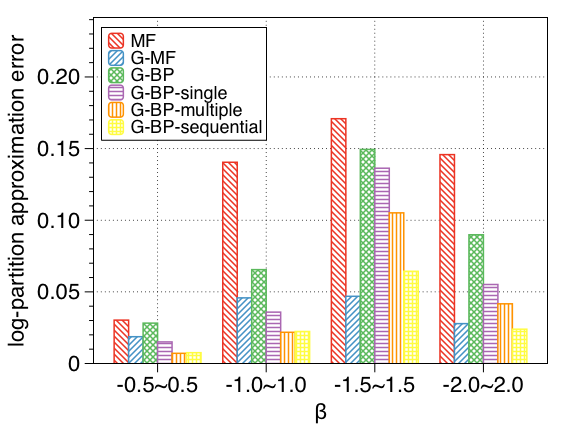}
    \end{minipage}
    \begin{minipage}{0.32\textwidth}
    \includegraphics[width=0.99\textwidth]{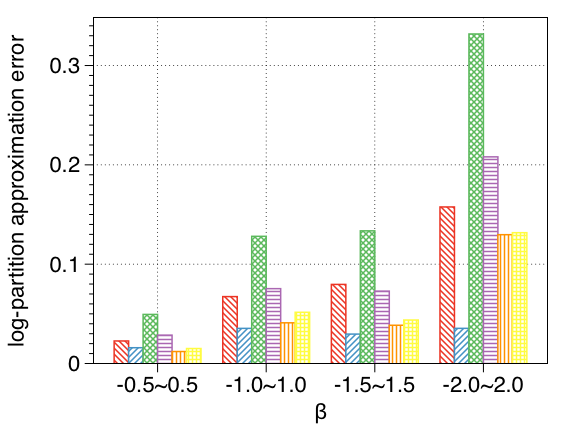}
    \end{minipage}
    \begin{minipage}{0.32\textwidth}
    \includegraphics[width=0.99\textwidth]{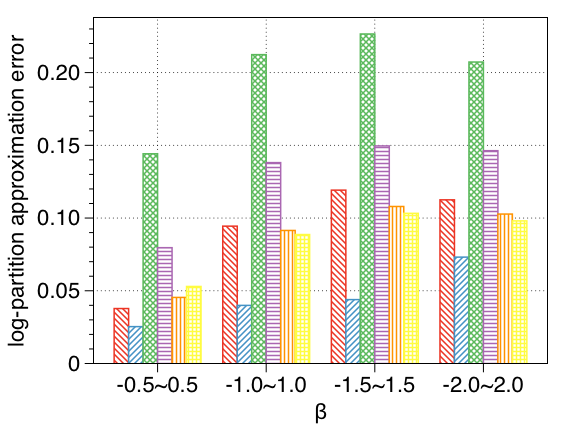}        
    \end{minipage}
    \caption{Averaged log-partition approximation error vs {interaction strength} 
    $\beta$ 
    in the case of generic (non-log-supermodular) GMs on complete graphs of size 4, 5 and 6 (left, middle, right), where the average is taken over 20 random
    models. 
    }
    \label{fig:exp1}
        \vspace{-0.1in}
\end{figure*}

\begin{figure*}[t!]
\centering
    \begin{minipage}{0.32\textwidth}
    \includegraphics[width=0.99\textwidth]{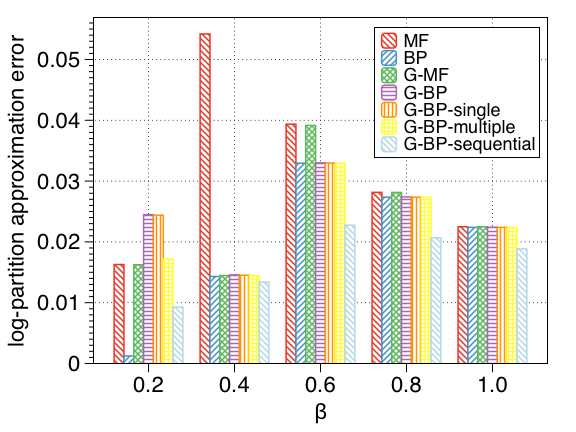}
    \end{minipage}
    \begin{minipage}{0.32\textwidth}
    \includegraphics[width=0.99\textwidth]{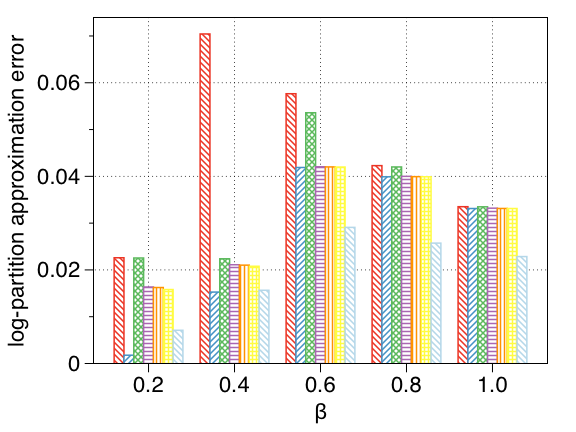}
    \end{minipage}
    \begin{minipage}{0.32\textwidth}
    \includegraphics[width=0.99\textwidth]{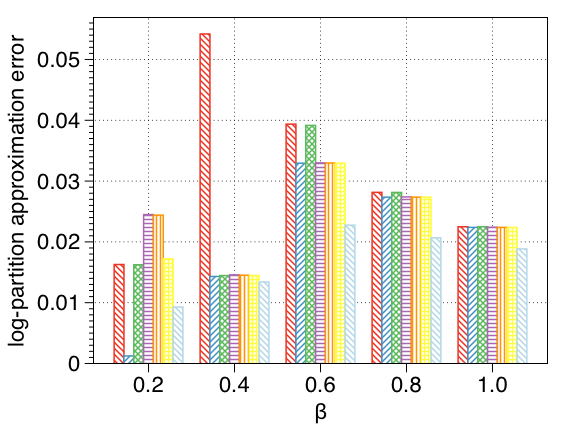}        
    \end{minipage}
    \caption{Averaged log-partition approximation error 
    vs {interaction strength} 
    $\beta$ 
    in the case of log-supermodular GMs on 
    complete graphs of size 4, 5 and 6 (left, middle, right), where the average is taken over 20 random
    models.}
    \label{fig:exp2}
    \vspace{-0.15in}
\end{figure*}

In this section, we report results of 
our experiments with G-MF and G-BP
defined in Section \ref{sec:alg}.
We also experiment here with G-BP boosted by schemes correcting errors by accounting for single \eqref{eq:fixerror1} and multiple \eqref{eq:fixerror2} terms, 
as well {as} correcting G-BP by applying G-BP sequentially again to each residual
partition function $Z_i$. The error decreases, while the evaluation complexity increases, as we move from G-BP-single to G-BP-multiple and then to G-BP-sequential.
As mentioned earlier, we use the IPOPT solver \citep{wachter2006implementation} to resolve the proposed gauge optimizations. We generate random GMs with factors dependent on 
the `interaction strength' parameters $\{\beta_{a}\}_{a\in \mathcal{V}}$ 
(akin inverse temperature) as follows:
\begin{equation*}
    \eff_{a}(x_{a}) = \exp(-\beta_{a}|h_{0}(x_{a}) - h_{1}(x_{a})|),
\end{equation*}
where $h_{0}$ and $h_{1}$ count numbers of $0$ and $1$ contributions in $x_{a}$,
respectively. Intuitively, we expect that as $|\beta_{a}|$ increases, it becomes more difficult to approximate the partition function. See the supplementary material for additional information on how we generate the random models. 

In the first set of experiments, we consider relatively small, complete graphs with two types of factors: random generic (non-log-supermodular) factors  and log-supermodular (positive/ferromagnetic) factors. 
Recall that the bare BP also provides a lower bound 
in the log-supermodular case \cite{ruozzi2012bethe}, thus making the comparison between each proposed algorithm and BP
informative.
We use the log partition approximation error 
defined as
$|\log Z - \log Z_{\text{LB}}|/|\log Z|$, 
where $Z_{\text{LB}}$ is the algorithm output (a lower bound of $Z$), to quantify the  algorithm's performance.
In the first set of experiments, we deal with relatively small graphs 
and the explicit computation of $Z$ (i.e., the approximation error)
is feasible. The results for experiments over the small graphs are illustrated in Figure~\ref{fig:exp1} and Figure~\ref{fig:exp2} for the non-log-supermodular and log-supermodular cases, respectively. Figure~\ref{fig:exp1} shows that, as expected, G-MF always outperforms MF. Moreover, we observe that G-MF typically provides the tightest low-bound, unless it is outperformed by G-BP-multiple or G-BP-sequential. 
We remark that BP is not shown in Figure~\ref{fig:exp1}, because in this non-log-supermodular case, it does not provide a lower bound in general. 
According to Figure \ref{fig:exp2}, showing the log-supermodular case, both G-MF and G-BP outperform MF, while G-BP-sequential outperforms all other algorithms. Notice that G-BP performs rather similar to BP in the log-supermodular case, thus suggesting that the constraints, distinguishing \eqref{eq:nnbetheopt} from \eqref{eq:betheopt}, 
are very mildly violated.

\begin{figure*}[t!]
    \vspace{-0.1in}
\centering
    \begin{minipage}{0.32\textwidth}
    \centering
    \includegraphics[width=0.99\textwidth]{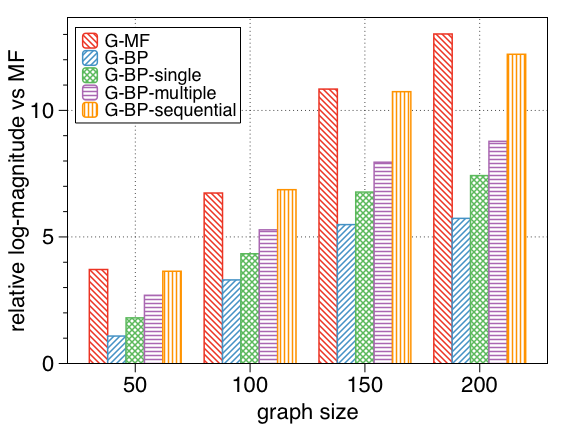}
    \end{minipage}
    \hspace{0.5in}
    \begin{minipage}{0.32\textwidth}
    \centering
    \includegraphics[width=0.99\textwidth]{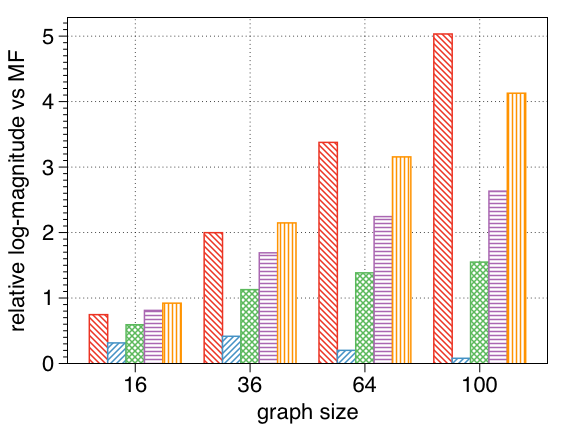}
    \end{minipage}
    \caption{Averaged ratio of the log partition function compared to MF 
    vs graph size (i.e., number of factors) in the case of generic (non-log-supermodular) GMs on 3-regular graphs (left) and 
    grid graphs (right), 
    where the average is taken over 20 random models.
  }
    \label{fig:exp3}
        \vspace{-0.1in}
\end{figure*}
\begin{figure*}[t!]
\centering
    \begin{minipage}{0.32\textwidth}
    \centering
    \includegraphics[width=0.99\textwidth]{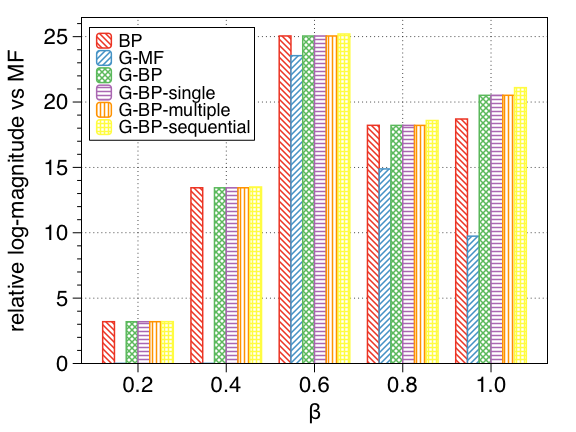}
    \end{minipage}
    \hspace{0.5in}
    \begin{minipage}{0.32\textwidth}
    \centering
    \includegraphics[width=0.99\textwidth]{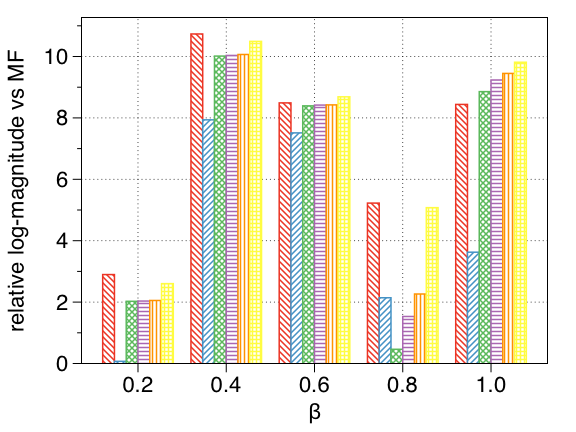}
    \end{minipage}
    \caption{Averaged ratio of the log partition function compared to MF 
    vs interaction strength 
    $\beta$ in the case of log-supermodular GMs on 3-regular graphs of size 200 (left) and 
    grid graphs of size 100 (right), 
    where the average is taken over 20 random models.
    }
    \label{fig:exp4}
        \vspace{-0.15in}
\end{figure*}

In the second set of experiments, we consider more sparse, larger
graphs of two types: $3$-regular and grid graphs with size up to 
$200$ factors/$300$ variables. 
As in the first set of experiments, the same non-log-supermodular/log-supermodular factors are considered. 
Since computing the exact approximation error is not feasible for the large graphs, we instead measure here the ratio of estimation
by the proposed algorithm to that of MF, i.e., 
$\log(Z_{\text{LB}}/Z_{\text{MF}})$
where $Z_{\text{MF}}$ is the output of 
MF. Note that a larger value of the ratio indicates better performance. 
The results are reported in Figure~\ref{fig:exp3} and Figure~\ref{fig:exp4}
for the non-log-supermodular and log-supermodular cases, respectively.
In Figure \ref{fig:exp3}, 
we observe that 
G-MF {and G-BP-sequential} outperform 
MF significantly,
e.g., up-to $e^{14}$ times 
better in $3$-regular 
graphs of size $200$.
We also observe that even the bare G-BP
outperforms MF. 
In Figure~\ref{fig:exp4}, 
algorithms associated with G-BP 
outperform G-MF and MF (up to $e^{25}$ times).
This is because 
the choice of $q(x)$ for G-BP
is favored by log-supermodular models, i.e.,
most of configurations are 
concentrated around $(0,0,\cdots)$ similar to the choice \eqref{eq:qchoice} of $q(x)$ for G-BP.
One observes here (again) that performance of G-BP in this log-supermodular case is almost on par with BP. This implies that G-BP generalizes BP well:
the former provides a lower bound of $Z$ for any GMs, 
while the latter does only for log-supermodular GMs.

\vspace{-0.1in}
\section{Conclusion and future research}

We explore the freedom in gauge transformations of GM
and develop novel variational inference methods which result in significant improvement of the partition function estimation.
In this paper, we have focused solely on designing approaches which improve the bare/basic MF and BP via specially optimized gauge transformations.
In terms of the path forward, it is of interest to extend this GT framework/approach to other variational methods, e.g., Kikuchi approximation \cite{kikuchi1951theory},
structured/conditional MF \cite{saul1996exploiting,carbonetto2007conditional}. 
Furthermore, G-BP and G-MF
were resolved in our experiments via
a generic 
optimization solver (IPOPT), which
was sufficient for the illustrative tests conducted so far,  however we expect that it might be possible to
develop more efficient distributed solvers of the BP-type. 
Finally, we plan working on applications of the newly designed methods and algorithms to a variety of practical inference applications associated to GMs.

\newpage
\appendix
\section{Construction of Forney-style model equivalent to factor-graph model}
\begin{figure*}[h!]
\centering
    \begin{minipage}{0.35\textwidth}
    \includegraphics[width=0.99\textwidth]{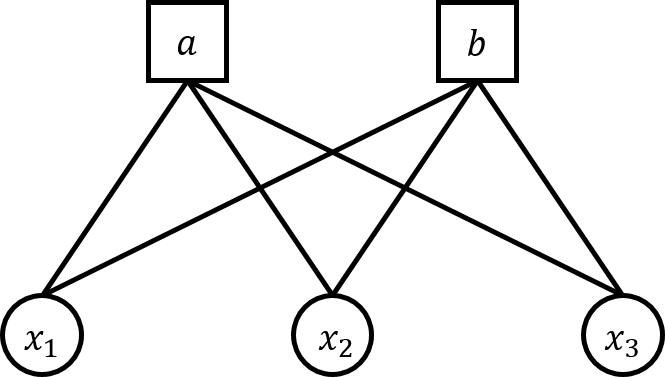}
    \end{minipage}
    \hspace{0.5in}
    \begin{minipage}{0.35\textwidth}
    \includegraphics[width=0.99\textwidth]{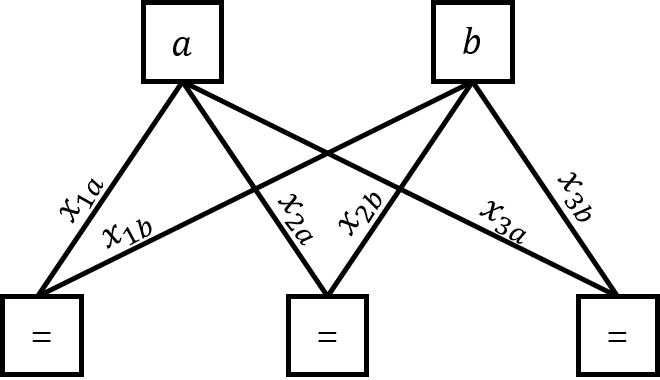}
    \end{minipage}
    \caption{Example of the transformation from the factor-graph GM (left) to the Forney-style GM (right). 
    Factors denoted as `=' constrains adjoining variables to have the same value. 
    Originally, the factor-graph GM had $3$ variables ($x_{1},x_{2},x_{3}$) and 2 factors ($a, b$). 
    In the equivalent Forney-style GM, there are $6$ variables ($x_{1a}, x_{1b}, x_{2a}, x_{2b}, x_{3a}, x_{3b}$) and $5$ factors ($a,b$ and three `=' factors).}
    \label{fig:factor2forney}
\end{figure*}

In this Section, we describe construction of a Forney-style GM equivalent to the factor-graph GM.
Consider a factor-graph GM defined on graph $G=(\mathcal{X}, \mathcal{F}, \mathcal{E})$ with factors $\{\eff_{a}\}_{a\in\mathcal{F}}$. 
Then one introduces the following Forney-style GM defined 
over the graph $(\mathcal{V},\mathcal{E})$ with
factors $\{\eff^{\dagger}_{a}\}_{a\in\mathcal{V}}$ 
\begin{align*}
    \mathcal{V} &\leftarrow \mathcal{X}\cup\mathcal{F},
    \qquad
    \eff^{\dagger}_{a} \leftarrow \eff_{a}, \quad \forall a \in \mathcal{F},\\
    \eff^{\dagger}_{a}(x_{a}) &\leftarrow
    \begin{cases}
    1 \quad \text{if }x_{a} = (1,1,\cdots)~\text{or}~(0,0,\cdots)\\
    0 \quad \text{otherwise}
    \end{cases},
    ~ \forall a\in\mathcal{X}.
\end{align*}
One observes that if the factor-graph GM (possibly, of high-order) is sparse, i.e., the maximum degree of $(\mathcal{X},\mathcal{F},\mathcal{E})$ is small,
then the equivalent Forney-style GM is too. See Figure \ref{fig:factor2forney} for illustration. 

\section{Proof of Theorem \ref{thm:optcycle}} 

To prove Theorem \ref{thm:optcycle} one, first, shows that 
the line graph GM can be gauge transformed into a distribution equivalent to 
the alternating cycle GM. 
Then it is sufficient for proving  Theorem \ref{thm:optcycle} to consider only the case of an alternating cycle. 

Consider  a GM defined on a line graph 
$G=(\mathcal{V},\mathcal{E})$
with $\mathcal{V} = \{a_{1},a_{2},\cdots,a_{n}\}$ and 
edges $\mathcal{E} = \{\{a_{1},a_{2}\},\{a_{2},a_{3}\},\cdots,\{a_{n-1},a_{n}\}\}$.
Then the gauge transformed
factor $\eff_{a_{i},\mathcal{G}}$
can be expressed as:
\begin{equation*}
\eff_{a_{i},\mathcal{G}} =
G_{a_{i}a_{i-1}}^{\top}
\eff_{a_{i}}
G_{a_{i}a_{i+1}},
\end{equation*}
where we used the fact that the size/cardinality of the factor is $2$.
Next, we `flip' factor $\eff_{2}$, associated with the node number $2$, such that there exist an 
odd number of negative definite factors among $\eff_{2},\cdots \eff_{n-1}$,
i.e.,
the flipping  sets
\begin{equation}
\label{eq:flipgauge}
G_{a_{1}a_{2}}, G_{a_{2}a_{1}}=
\left[
\begin{array}{ccc}
0 & 1 \\
1 & 0 \\
\end{array}
\right],
\end{equation}
thus resulting  in reversing the sign of $\text{det}(\eff_{a_{2}})$.
If $\eff_{a_{2}}$ is non-invertible, i.e. $\text{det}(\eff_{a_{2}}) = 0$, we instead flip $\eff_{3}$ and so on. 
If all  factors are non-invertible, the resulting distribution is 
a product distribution and one can easily 
find the optimal $q$ for the corresponding line graph, which completes the proof. Otherwise, we `join' the endpoints $a_{1}, a_{n}$ into $a_{0}$ by
introducing a non-invertible factor $\eff_{0} = \eff_{1}\eff_{n}^{\top}$,
which results in an alternating cycle with the probability distribution identical to the one of a line graph GM.

Our next step is to prove Theorem \ref{thm:optcycle} for an alternating cycle GM. Our high level logic here is as follows. We first fix the distribution $q$ of \eqref{eq:gibbsopt} according to 
\begin{equation*}
\begin{split}
    q(x) = 
    \begin{cases}
    1 \qquad &\text{if } x = (0,0,\cdots),\\
    0 \qquad &\text{otherwise}.
    \end{cases},
\end{split}
\end{equation*} 
and then show that the GM can be gauge transformed into a 
distribution with a nonzero probability concentrated only at $(0,0,\cdots)$. The resulting objective of \eqref{eq:gibbsopt} 
will become exactly the partition function. 
To implement this logic, consider an alternating cycle
defined on some graph $G=(\mathcal{V},\mathcal{E})$
with $\mathcal{V} = \{a_{1},a_{2},\cdots,a_{n}\}$ and
edges $\mathcal{E} =
\{\{a_{1},a_{2}\},\{a_{2},a_{3}\},\cdots,\{a_{n-1},a_{n}\}, \{a_{n}, a_{1}\}\}$.
Observe that, that the gauge transformed factor,
$\prod_{i}\eff_{a_{i},\mathcal{G}}$, and the original factor,
$\prod_{i}\eff_{a_{i}}$,
share a pair of eigenvalues
$\lambda_{1}, \lambda_{2}$ due to the following relationship: 
\begin{equation*}
\prod_{i}\eff_{a_{i},\mathcal{G}} =
G_{a_{n}a_{1}}^{-1}\prod_{i}\eff_{i}G_{a_{n}a_{1}}
\end{equation*}
One finds that
$\lambda_{1}\lambda_{2} = \prod\text{det}(\eff_{i}) \leq 0$
since there exist an odd number of negative definite factors in
the cycle.
Moreover,
$\lambda_{1} + \lambda_{2} > 0$
because the diagonal sum,
$\prod_{i}\eff_{i}$,
is equivalent to
the partition function of GM.
Thus one can assume, without loss of generality, that $\lambda_{1}>0$ and $\lambda_{2}<0$.

Next, utilizing a simple linear algebra, one derives
\begin{equation*}
Q_{2}^{-1}Q_{1}G_{a_{n}a_{1}}\prod_{i}\eff_{i,\mathcal{G}}Q^{-1}_{1}Q_{2}
= \left[\begin{array}{ccc}
\lambda_{1}+\lambda_{2} & \lambda_{1} \\
-\lambda_{2} & 0 \\
\end{array}
\right],
\end{equation*}
where $Q_{1}$ and $Q_{2}$ are matrices whose $j$-th
column is an eigen-vector of $\prod_{i}\eff_{i}$
and,
$\left[\begin{array}{ccc}
\lambda_{1}+\lambda_{2} & \lambda_{1} \\
-\lambda_{2} & 0 \\
\end{array}
\right]$,
respectively.
Now let
\begin{align*}
G_{a_{n}a_{1}} = Q_{1}^{-1}Q_{2}, \qquad
G_{a_{i-1}a_{i}} = (\eff_{i}G_{a_{i}a_{i+1}}^{\top})^{-1}
\quad\text{for}\quad i=2,\cdots n,
\end{align*}
where $a_{n+1}=a_{1}$.
Here we assume that there exists at most
one non-invertible factor in the GM and
$\eff_{2},\cdots,\eff_{n}$ are
invertible so that $(\eff_{i}G_{a_{i}a_{i+1}}^{\top})^{-1}$
is defined properly.
Otherwise,
the GM can be decomposed
into separate line graphs and the
proof can be applied recursively.
Then the gauge transformed factors become:
\begin{align*}
\eff_{a_{1},\mathcal{G}} =
\left[
\begin{array}{ccc}
\lambda_{1}+\lambda_{2} & \lambda_{1} \\
-\lambda_{2} & 0 \\
\end{array}
\right],
\quad
\eff_{a_{i},\mathcal{G}} =
\left[
\begin{array}{ccc}
1 & 0 \\
0 & 1 \\
\end{array}
\right]\quad \forall i \neq 1,
\end{align*}
which corresponds to a GM with objective of \eqref{eq:gibbsopt}
to be equal to the log partition function.
This completes the proof of the Theorem \ref{thm:optcycle}.

\section{Generating GM instances (for experiments)}
In this Section, we provide more details on our experimental setups reported in
in Section \ref{sec:experiments}. 
First, we explain how the two types 
of factors, non-log-supermodular and log-supermodular, were constructed. 
In the generic case (of non-log-supermodular factors), i.e., correspondent to Figure \ref{fig:exp1} and Figure \ref{fig:exp3}, 
one generates factor by first drawing the interaction strength vector at random from the i.i.d. 
uniform distribution over the interval $[-T,T]$ for some $T>0$, i.e.,
$\beta_{a} \sim \mathcal{U}(-T, T)$.
Then, in order to introduce  a bias, 
we add an external variable $y_{a}$, i.e., half-edge, as follows:
\begin{equation*}
    \eff_{a}(x_{a}) = \exp(\beta_{a}|h_{0}(x_{a}\cup y_{a}) - h_{1}(x_{a} \cup y_{a})|), 
\end{equation*}
where $y_{a}$ is either 
$\{0\}$ or $\{1\}$ with probability $1/2$ each. 
More specifically in experiments resulted in Figure \ref{fig:exp1} one varies $T$ while in the experiments resulted in Figure \ref{fig:exp3} one fixes $T$ to $1.0$, i.e., $\beta_{a} \sim [-1.0, 1.0]$.
Next, in the case of the log-supermodular factors, i.e., setting resulted in Figure \ref{fig:exp2} and Figure \ref{fig:exp4}, one generates log-supermodular factors 
by drawing the interaction strength vector from 
normal distribution with the average $T>0$ and the variance, $10^{-4}$, i.e., $\beta_{a} \sim \mathcal{N}(T, 10^{-4})$. 
Note that there exist no bias in the factors and even though the distribution of the interaction strength is normal, it is highly likely to observe  a positive value concentrated around $T$. 

\end{document}